\documentclass[letterpaper]{article} 
\usepackage{aaai2026}  
\usepackage{times}  
\usepackage{helvet}  
\usepackage{courier}  
\usepackage[hyphens]{url}  
\usepackage{graphicx} 
\usepackage{natbib}  
\usepackage{caption} 
\usepackage{algorithm}
\usepackage{algorithmic}
\usepackage{newfloat}
\usepackage{listings}
\DeclareCaptionStyle{ruled}{labelfont=normalfont,labelsep=colon,strut=off} 
\floatstyle{ruled}
\newfloat{listing}{tb}{lst}{}
\floatname{listing}{Listing}
\usepackage{amssymb}
\usepackage{amsmath}
\usepackage{subfigure}

\title{ConceptFlow: Hierarchical and Fine-grained Concept-Based Explanation for Convolutional Neural Networks}

\author{
    Xinyu Mu\textsuperscript{\rm 1,2},
    Hui Dou\textsuperscript{\rm 1,4},
    Furao Shen\textsuperscript{\rm 1,4}\thanks{Corresponding author: frshen@nju.edu.cn}
    Jian Zhao\textsuperscript{\rm 3},
}
\affiliations{
    \textsuperscript{\rm 1}National Key Laboratory for Novel Software Technology, Nanjing University, China \\
    \textsuperscript{\rm 2}School of Artificial Intelligence, Nanjing University, China \\
    \textsuperscript{\rm 3}School of Electronic Science and Engineering, Nanjing University, China \\
    \textsuperscript{\rm 4}School of Computer Science and Technology, Nanjing University, China \\
    huidou@smail.nju.edu.cn, frshen@nju.edu.cn
}

\usepackage{bibentry}

\begin{document}

\maketitle

\begin{abstract}



Concept-based interpretability for Convolutional Neural Networks (CNNs) aims to align internal model representations with high-level semantic concepts, but existing approaches largely overlook the semantic roles of individual filters and the dynamic propagation of concepts across layers. To address these limitations, we propose ConceptFlow, a concept-based interpretability framework that simulates the internal ``thinking path” of a model by tracing how concepts emerge and evolve across layers. ConceptFlow comprises two key components: (i) concept attentions, which associate each filter with relevant high-level concepts to enable localized semantic interpretation, and (ii) conceptual pathways, derived from a concept transition matrix that quantifies how concepts propagate and transform between filters. Together, these components offer a unified and structured view of internal model reasoning.
Experimental results demonstrate that ConceptFlow yields semantically meaningful insights into model reasoning, validating the effectiveness of concept attentions and conceptual pathways in explaining decision behavior.
By modeling hierarchical conceptual pathways, ConceptFlow provides deeper insight into the internal logic of CNNs and supports the generation of more faithful and human-aligned explanations.

\end{abstract}


\section{Introduction}

Owing to their exceptional performance across a wide range of tasks, Convolutional Neural Networks (CNNs) have seen widespread adoption, thereby intensifying the demand for improved interpretability \cite{DBLP:journals/tmi/WenYLCX25,DBLP:conf/bibm/WuCCR23,DBLP:conf/cvpr/PatricioNT23}.
A variety of methods have been developed to interpret black box models, primarily aiming to generate explanations by identifying the key features the model relies on for its predictions \cite{DBLP:conf/cvpr/CheferGW21,DBLP:journals/tvcg/NetoP21,DBLP:journals/tvcg/GolbiowskaC22,DBLP:series/lncs/KindermansHAASDEK19,DBLP:journals/tec/SuVS19,DBLP:journals/tmm/ZhangLLGLXZ25,DBLP:conf/cvpr/00030K23}. 
Although these techniques are widely adopted, they have faced criticism for relying on low-level inputs such as pixel intensities or raw sensor data, which are often combined in opaque ways and fail to align with high-level, semantically meaningful concepts that are intuitive to humans \cite{DBLP:conf/cvpr/RamaswamyKFR23,DBLP:conf/icml/ChoiRF0J023}.

Among various interpretability paradigms, concept-based explanation methods have received significant attention for their potential to align machine reasoning with human understanding. 
By grounding model behavior in high-level human-interpretable concepts, these methods aim to make model predictions more transparent and trustworthy \cite{DBLP:conf/aaai/00010025,DBLP:conf/cvpr/0004HTYX25,DBLP:journals/tai/DoumanoglouAZ24,DBLP:conf/cvpr/DreyerASL22,DBLP:conf/eccv/LeeMSWW24}.
Most existing concept-based approaches focus on identifying and attributing concepts at specific layers or decisions, overlooking two critical aspects: (1) the association of individual filters with meaningful concepts, (2) the dynamic propagation and abstraction of hierarchical concepts throughout the model.
For instance, a concept learned at one filter in a layer (e.g., "cornered shape") transforms into another concept in the filter in the subsequent layer (e.g., "digit 7"). 
This process mirrors the way humans build reasoning chains through successive layers of understanding. 
Modeling such conceptual transformations provides deeper insights into the internal logic of neural networks and facilitates the generation of human-aligned explanations.

\begin{figure}
    \centering
    \includegraphics[width=0.8\linewidth]{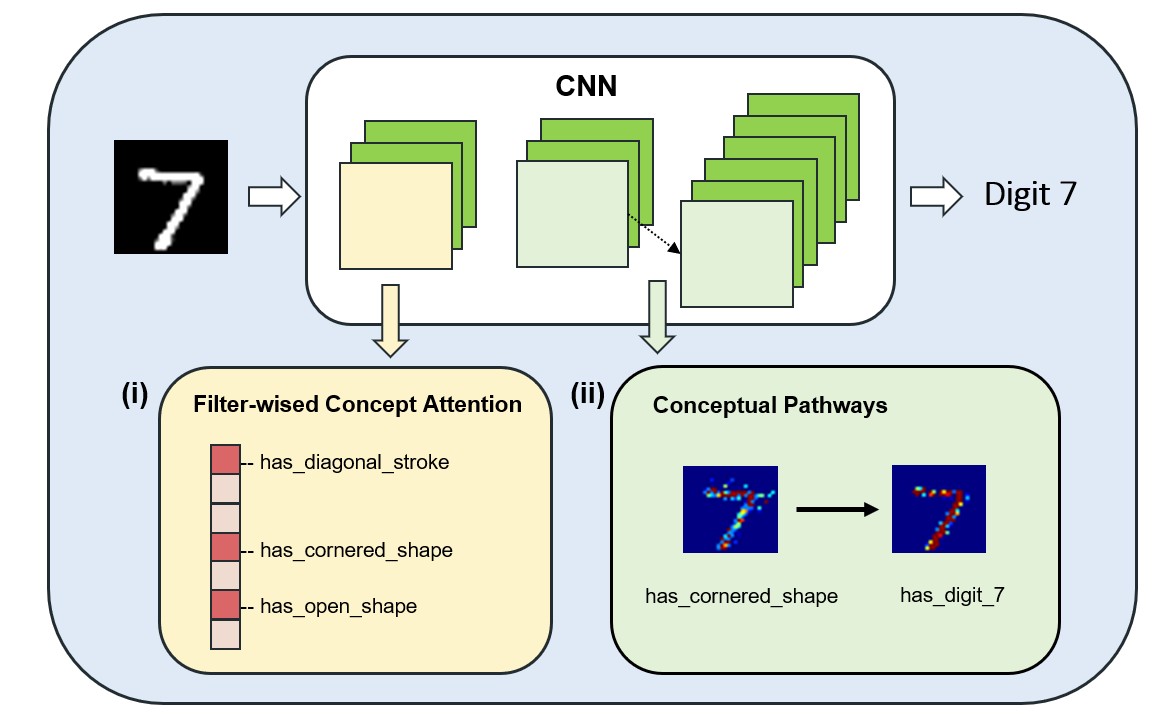}
    \caption{The framework of ConceptFlow. It (i) generates concept attention for each filter, highlighting the relevant concepts, and (ii) extracts conceptual pathways between layers, revealing how concepts propagate within the network.
}
    \label{fig:display}
\end{figure}

In this paper, we propose the \textit{ConceptFlow}, a concept-based interpretability framework that simulates the network’s internal ``thinking path” by tracing the flow of hierarchical semantic information across layers. As shown in Figure~\ref{fig:display}, the framework consists of two key components: (i) \textit{concept attention}, which associates each filter to a set of concepts, enabling localized semantic grounding, and (ii) \textit{conceptual pathway}, which captures how hierarchical concepts propagate and transform between filters across layers.
Together, these components offer a unified interpretability framework that captures both the static alignment between filters and concepts, and the dynamic flow of concepts across layers, enabling a structured and principled analysis of the model’s internal reasoning process.

Our main contributions are summarized as follows:
\begin{itemize}
\item Generation of filter-wise concept attentions. ConceptFlow introduces a mechanism that associates each filter with relevant high-level human-interpretable concepts. This is achieved by extending attention from low-level input features to abstract concepts, enabling fine-grained, localized interpretation of filter semantics.

\item Extraction of conceptual pathways. The concept transition matrix is proposed to quantify the transformation of concepts between layers. This matrix quantitatively characterizes how concepts propagate and transform within the network hierarchically, enabling a layer-wise analysis of the internal semantic flow.

\item Experiments demonstrate that ConceptFlow provides meaningful insights into model reasoning. Results confirm the effectiveness of concept attentions and show that conceptual pathways correlate with key decision-making patterns, highlighting their importance for understanding model behavior.
\end{itemize}

This work presents a new direction for concept-based interpretability, emphasizing not just what concepts are present, but how they are formed and transformed throughout the model.

\section{Related Work}
Concept-based interpretability methods aim to enhance the transparency of deep neural networks by linking internal representations with high-level, human-understandable concepts. 
Testing with Concept Activation Vectors (TCAV) \cite{kim2018interpretability} offers an alternative paradigm that does not require architectural modifications or retraining. TCAV defines user-specified concepts using examples and trains linear classifiers to extract concept activation vectors from internal representations.
Concept Bottleneck Models (CBM) \cite{koh2020concept} and Logic Explained Networks (LEN) \cite{ciravegna2023logic,barbiero2022entropy,DBLP:conf/emnlp/JainCBGBL22} encode concepts either through dedicated neurons or low-dimensional embeddings. 
These approaches involve explicitly supervising intermediate layers with concept annotations that describe semantic attributes relevant to the target classes \cite{DBLP:conf/icml/BarbieroCGZMTLP23,DBLP:conf/nips/ZarlengaBCMGDSP22,DBLP:conf/icml/KimJPKY23,DBLP:conf/cvpr/0004HTYX25,DBLP:conf/cvpr/LiuZG25a,DBLP:conf/aaai/HuangSHZWS24,DBLP:conf/icml/KimJPKY23,DBLP:conf/icml/ShinJAL23}. 
When such annotations are available in the training dataset, the model can be jointly optimized for both the primary task (such as image classification) and concept learning. 
These models facilitate analysis of class-concept relationships by enabling concept attribution, importance scoring, or logic-based reasoning.

When concept annotations are not available in the primary training dataset, concept instillation techniques provide a mechanism to incorporate semantic information from external sources \cite{DBLP:conf/iclr/RigottiMGGS22,DBLP:journals/natmi/ChenBR20,DBLP:conf/cikm/KazhdanDJLW20,DBLP:conf/iclr/YuksekgonulW023,DBLP:conf/cvpr/KowalDAGDT24}. 
These methods typically follow a two-stage process: first training the model on the main task, and then aligning internal representations with concepts using auxiliary resources. 
The methods leverage external labeled datasets, semi-supervised corpora, or structured knowledge graphs to embed concept structure into the model \cite{DBLP:conf/aaai/RanWCGZ0S25,DBLP:journals/tmi/WenYLCX25,DBLP:journals/tgrs/ZhangRHMLJ25,DBLP:conf/aaai/SpeerCH17,DBLP:journals/pr/XuHZCSC25}. 
In some variants, the final classifier is replaced with an interpretable predictor to support downstream concept reasoning.

These methods treat concepts as isolated and static elements associated with specific layers. The progression and interaction of concepts across multiple layers remain insufficiently explored. A deeper understanding of how concepts evolve hierarchically within models is essential for developing more comprehensive interpretability frameworks.

\section{Method}

\begin{figure*}
    \centering
    \includegraphics[width=0.8\linewidth]{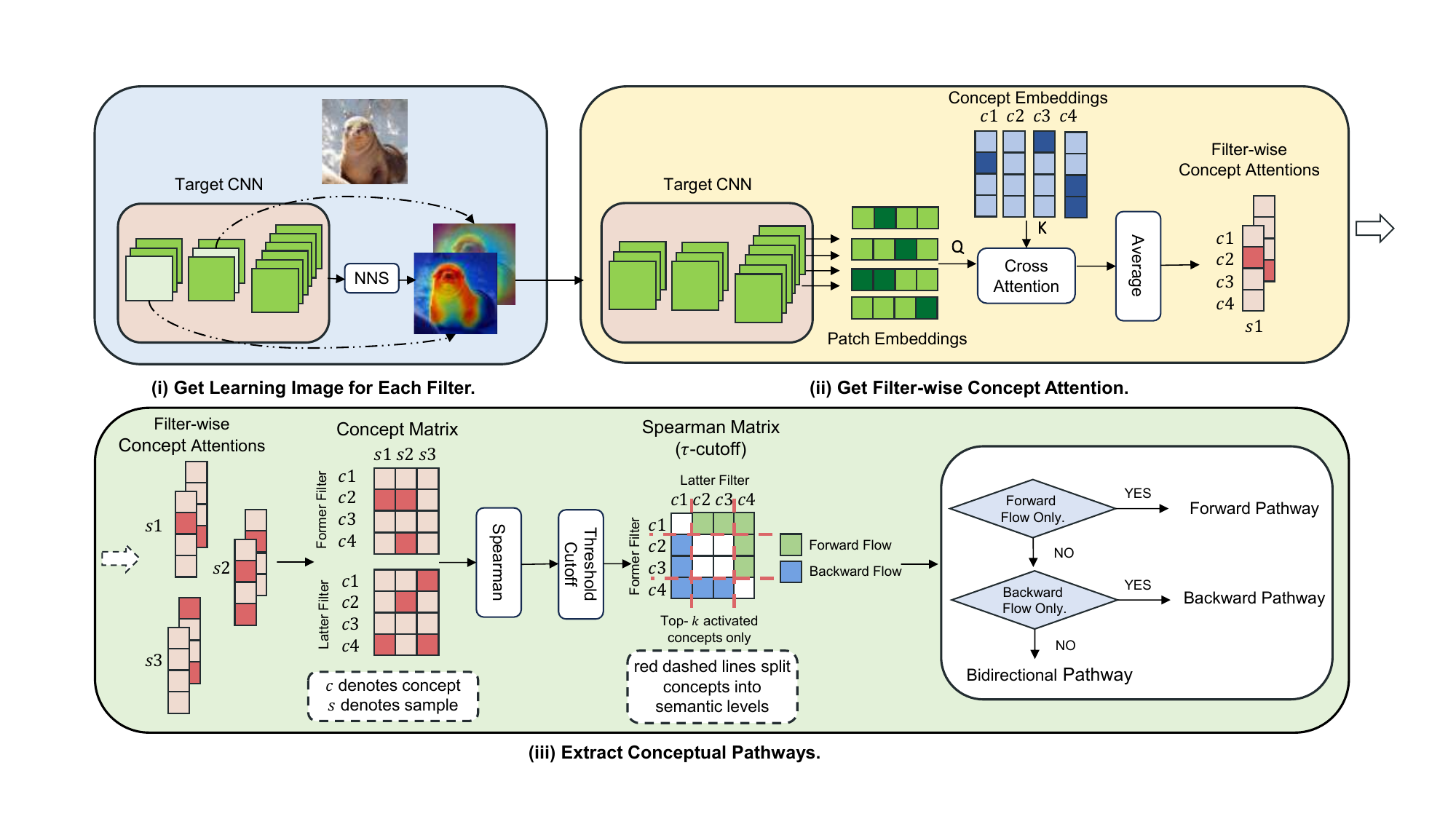}
    \caption{The overall procedure of ConceptFlow applied to two filters in adjacent layers. (i) Use the Neural Network Scanner (NNS) to generate learning images that visualize the patterns captured by each filter.
    (ii) Apply these learning images to the original input, producing filter-wise concept attentions that highlight the semantic focus of individual filters.
    (iii) Concatenate the concept attentions across multiple samples. Use Spearman correlation to extract the conceptual pathways by analyzing monotonic relationships between concepts. Conceptual flows are further categorized into forward/backward based on the relative semantic levels of involved concepts.}
    \label{fig:procedure}
\end{figure*}

We propose ConceptFlow, a concept-based interpretability framework for CNNs that explains how concepts emerge and evolve across layers. The overall procedure is illustrated in Figure~\ref{fig:procedure}. The process begins with computing filter-wise concept attentions, which offers fine-grained semantic interpretation of individual filters. This is achieved by applying filter-wise learning images to the original inputs and feeding the resulting representations into the cross-attention module. Subsequently, conceptual pathways are extracted by analyzing monotonic relationships between concepts. They characterize how concepts propagate and transform across layers. Finally, we construct the concept transition matrix to formalize transformation of concepts.

\subsection{Preliminary}

Our method is built upon two key components: the Concept Transformer (CT) \cite{rigotti_attention-based_2022} and the Neural Network Scanner (NNS) \cite{dou_understanding_2023}.

\textbf{Concept Transformer.} The Concept Transformer is a drop-in replacement for the classifier head in standard deep learning models. It computes the cross attention between the input features and a predefined set of domain-specific concepts. Given an image $\mathbf{I}$ and a concept set $\mathcal{C} = \{c_1, \ldots, c_{n_c}\}$ composed of $n_c$ concepts, it outputs an attention vector $\mathbf{a} \in \mathbb{R}^{n_c}$, where each element $(\mathbf{a})_i$ reflects the relevance of concept $c_i$ to the input. The vector $\mathbf{a}$ is referred to as the \emph{concept attention}, which serves as a foundation for our analysis of conceptual pathways.

\textbf{Neural Network Scanner.} The Neural Network Scanner (NNS) is a post-hoc visualization technique designed to reveal the internal representations learned by neural networks. For CNNs, it generates a \emph{learning image} $\mathbf{I_L}$ for each convolutional filter, which maintains the same spatial resolution as the input. These learning images highlight the feature patterns captured by each filter. In our method, NNS is used to mediate attention allocation, directing focus toward the concepts encoded within convolutional filters.

\subsection{Obtaining Filter-Wise Concept Attentions}
Initially, we provide fine-grained semantic interpretations for individual filters by computing their corresponding concept attentions, which associates each filter to relevant human-understandable concepts through attention scores. Consider the CNN with $L$ layers, where $F^m_l$ stands for the $m$-th filter in layer $l$. Given an original input $\mathbf{I} \in \mathbb{R}^{C \times H \times W}$, we apply the NNS to generate a learning image for each filter $F^m_l$. This learning image is then overlaid to the original input with a blending ratio of 1, resulting in a modified input denoted as $\mathbf{I}^m_l$, which preserves the original spatial resolution. Since the learning image reflects the learned features, blending it with the original input enhances the specific concept representations encoded within the filter. 

We input each $\mathbf{I}^m_l$ to the CNN backbone directly, which yields the corresponding feature map $\mathbf{M}^m_l \in \mathbb{R}^{C^{\prime} \times H^{\prime} \times W^{\prime}}$. The feature map is then divided into $P$ visual patches. Each visual patch is projected to a query vector $\in \mathbb{R}^{d}$, forming a query matrix $\mathbf{Q}^m_l \in \mathbb{R}^{P \times d}$. Given a predefined concept set $\mathcal{C} = \{c_1, \ldots, c_{n_c}\}$ consisting of $n_c$ concepts with hierarchical structure, we construct a trainable key matrix $\mathbf{K} \in \mathbb{R}^{n_c \times d}$. The cross attention matrix between the input features and concept embeddings can be written as follows:
\begin{equation}
    \mathbf{O}^m_l=\text{softmax}(\frac{1}{\sqrt{d}}\mathbf{Q}^m_l\mathbf{K}^\top), \quad \text{where} \quad \mathbf{O}^m_l \in \mathbb{R}^{P \times n_c}
\end{equation}
We compute the column-wise average of $\mathbf{O}^m_l$, and output the concept attention $\mathbf{a}^m_l \in \mathbb{R}^{n_c}$:
\begin{equation}
    (\mathbf{a}^m_l)_i= \frac{1}{P}\sum_{p=1}^{P}(\mathbf{O}^m_l)_{pi},\quad \text{where} \quad i=1,\cdots,n_c
\end{equation}
In this way, we compute the concept attention $\mathbf{a}^m_l$ for each filter $F^m_l$ given the input image $\mathbf{I}$, where the $i$-th element indicates the relevance between $\mathbf{I}^m_l$ and concept $c_i$.

\subsection{Extracting Conceptual Pathways}\label{method:step2}
Beyond the static concept attentions of individual filters, we aim to uncover how concepts are dynamically propagated across layers. Therefore, we introduce the \emph{conceptual pathway} -- a logical trajectory composed of predefined concepts that reflects the hierarchical transformation and evolution of semantics across layers. This pathway mirrors human reasoning by abstracting and integrating concepts.

Initially, we aggregate filter-wise concept attentions across input samples. Consider there are $S$ input images in total. For filter $F^m_l$, we collect its concept attentions across all inputs to form the \emph{concept attention matrix}:
\begin{equation}
    \mathbf{A}^m_l=[(\mathbf{a}^m_l)_1^\top,\cdots,(\mathbf{a}^m_l)_S^\top] \in \mathbb{R}^{n_c \times S}
\end{equation}
where $(\mathbf{a}^m_l)_s$ denotes the concept attention of the $s$-th input image. Equivalently, $\mathbf{A}^m_l$ can be represented as:
\begin{equation}
    \mathbf{A}^m_l=[(\mathbf{b}^m_l)_1^\top,\cdots,(\mathbf{b}^m_l)_{n_c}^\top]^\top
\end{equation} 
where each row vector $(\mathbf{b}^m_l)_c \in \mathbb{R}^S$ represents the attention distribution of the $c$-th concept over all input images, capturing how the focus on this concept varies across inputs.

Concept attention distributions are further analyzed to extract the conceptual pathways. Intuitively, if concept A contributes to the emergence of concept B, filters in shallower layers should first capture features related to concept A, which are then propagated to deeper layers. As a result, the attention distributions of these two concepts are expected to exhibit similar trends across input images: when concept A is strongly activated, concept B is more likely to be learned; conversely, if concept A is suppressed, concept B do not emerge due to the absence of foundational representations.

We employ the Spearman’s rank correlation coefficient (Spearman coefficient) to quantify the monotonic relationship. Given two input vectors $\in \mathbb{R}^n$, they are ranked respectively and the Spearman coefficient $\rho$ is defined as:
\begin{equation}
    \rho = 1 - \frac{6 \sum_{i=1}^{n} d_i^2}{n(n^2 - 1)} \in [-1,1]
\end{equation}
where \( d_i \) denotes the rank difference of the $i$-th elements. A stronger monotonic relationship corresponds to an absolute Spearman coefficient closer to 1.

For any two filters in adjacent layers, we compute the absolute Spearman coefficient for each concept pair, resulting in a \emph{Spearman matrix} $\mathbf{P} \in \mathbb{R}^{n_c \times n_c}$. Let $\mathbf{P}_l^{m,n}$ denotes the Spearman matrix between the filter $F^m_l$ and $F^n_{l+1}$, which is computed as follows:  
\begin{align}
    (\mathbf{P}_l^{m,n})_{ij} &= |\text{spearman}((\mathbf{b}_l^m)_i,(\mathbf{b}_{l+1}^n)_j)| \notag \\
    & \text{where}\quad i,j=1,\cdots,n_c
\end{align}
A threshold $\tau$ is set to retain strong monotonic relationships. If $(\mathbf{P}_l^{m,n})_{ij} \geq \tau$, we extract the \emph{conceptual flow} from concept $i$ to concept $j$ between the filter $F^m_l$ and $F^n_{l+1}$. Given the hierarchical organization of the concept set, conceptual flows can be categorized based on the relative semantic levels of the concepts involved: if concept $i$ is at a higher level than concept $j$, the flow is considered \emph{backward}; otherwise, it is \emph{forward}. Accordingly, we classify the inter-filter conceptual pathways into three types: a pathway is labeled as a \emph{forward pathway} or \emph{backward pathway} if it consists solely of forward or backward flows respectively; if both types of flows are present, it is defined as a \emph{bidirectional pathway}.

To enhance the robustness of extracted conceptual pathways, we compute the Spearman coefficient only for the top $k$ most-activated concepts of each filter, denoted by the set \(\mathbf{K}^m_l\). This ensures that only strongly activated concepts are involved in the extracted pathways. Additionally, we apply importance sampling to mitigate bias from the original concept distribution of the dataset. Finally, we exclude self-transitions where the same concept appears on both ends of a pathway. Ablation studies (Supplementary Material 4.1) demonstrate the effectiveness of these selection criteria.

\subsection{Formalization via the Concept Transition Matrix}
While the thresholding strategy facilitates the extraction of conceptual pathways based on strong pairwise correlations, it remains heuristic. To achieve a more rigorous representation, we introduce the \emph{concept transition matrix}, which formalizes the transformation between concepts explicitly.

As mentioned in Section~\ref{method:step2}, conceptual pathways are more likely to form between concepts exhibiting similar monotonic relationships. Consequently, we derive the concept transition matrix $\mathbf{T}$ directly from the Spearman matrix $\mathbf{P}$ by row normalization:
\begin{equation}
    (\mathbf{T}_l^{m,n})_{ij} = \frac{(\mathbf{P}_l^{m,n})_{ij}+\epsilon}{\sum_{t=1}^{n_c}\left((\mathbf{P}_l^{m,n})_{it}+\epsilon\right)}
\end{equation}
We add a small $\epsilon$ to each element for numerical stability. The $(i,j)$-th entry of $\mathbf{T}_l^{m,n}$ denotes the probability of transitioning from concept $i$ to concept $j$ from the filter $F^m_l$ to $F^n_{l+1}$. We further define the \emph{concept transition mass} as the column-wise sum of the transition matrix, quantifying the total conceptual flow toward each downstream concept.

The concept transition matrix satisfies the following properties:

\textbf{Property 1. Local Heterogeneous Markov Property.}
Conceptual transitions between adjacent filters can be viewed as local Markov processes. Given filters $F^m_l$ and $F^n_{l+1}$, the concept distribution at $F^n_{l+1}$ depends only on that of $F^m_l$ via the transition matrix $\mathbf{T}_l^{m,n}$. Moreover, fixing a sequence of $n$ filters ($f_1,f_2,\cdots,f_n$) across $n$ layers yields an $n$-step heterogeneous Markov process, where the total transition is the product of the $n - 1$ local matrices:
\begin{equation}
    \mathbf{T}=\prod_{l=1}^{n-1} \mathbf{T}_l^{f_l,f_l+1}
\end{equation}

\textbf{Property 2. Faithfulness.}
Faithfulness is guaranteed when the interpretations indeed explain the model’s operation \cite{lakkaraju_interpretable_2017}. Specifically, the conceptual transition is considered \emph{faithful} since the involved concepts are genuinely activated and learned by the corresponding filters.  This property is ensured by selecting top $k$ concepts with largest activation as mentioned in Section~\ref{method:step2}: 
\begin{equation}
    (\mathbf{T}_l^{m,n})_{ij} \geq \tau_{\text{eff}} \Rightarrow i \in \mathbf{K}^m_l \land j \in \mathbf{K}^n_{l+1}
\end{equation}
where $\tau_{\text{eff}}$ is a small positive threshold that filters out uniform or spurious transitions introduced by $\epsilon$. 

\textbf{Property 3. Spectral-based Semantic Convergence.}
Given the concept transition matrix, the spectral radius of non-dominant eigenvalues controls the convergence speed of concept propagation. A small spectral radius implies rapid abstraction and semantic compression, while larger spectral radius preserves diverse conceptual modes.

Proof of the above properties is provided in Section 1 of Supplementary Material.

\section{Experiments}

We conduct various experiments to confirm the effectiveness of ConceptFlow. We first demonstrate that filter-wise concept attentions reflect localized semantics faithfully. Then we extract the inter-layer conceptual pathways, explaining how concepts evolve and transform. Finally, we validate the importance of conceptual pathways on the decision-making process of models. All experiments are performed on two self-annotated datasets with hierarchical concepts.


\textbf{Datasets.} To capture semantic transitions among concepts, we annotate two datasets with fine-grained, hierarchical concept labels. The first is \emph{Concept-MNIST (CMNIST)} built upon MNIST\cite{726791}. Instead of classifying digits from 0 to 9, we perform binary classification: even vs.\ odd. The concept set contains 21 concepts across 3 levels: (i) basic strokes (e.g., vertical, horizontal), (ii) local structures (e.g., open, cornered shapes) composed of strokes, and (iii) digit identities (0--9) as the final semantic level.


The second dataset is \emph{Concept-AwA-10 (CAWA)} which is derived from Animals with Attributes\cite{8413121}. We select 10 classes and perform binary classification: carnivorous vs.\ herbivorous. After removing inactive attributes and adding animal identities as high-level concepts, we construct a hierarchical concept set of 49 concepts across 4 levels. Concepts range from low-level features (e.g., color, texture), to body parts, then physical traits and habits, and finally animal identities. Full annotations of both datasets are provided in Section 2 of Supplementary Material.

\textbf{Models.} We adopt CNN backbones with the cross attention module for both datasets. The CNN consists exclusively of 3$\times$3 filters across all layers. A 3-layer and 5-layer CNN are employed respectively for CMNIST and CAWA. Each layer contains 64 filters, except the final layer which uses 128. All models are trained end-to-end (details in Section 3 of Supplementary Material), and achieve validation accuracy of 99.6\% and 83.9\% respectively.


\subsection{Interpreting Filter-Wise Semantics}\label{sec:local-level} 

\begin{figure}[htbp]
    \centering
    \subfigure[Original.]{
        \includegraphics[width=0.20\textwidth]{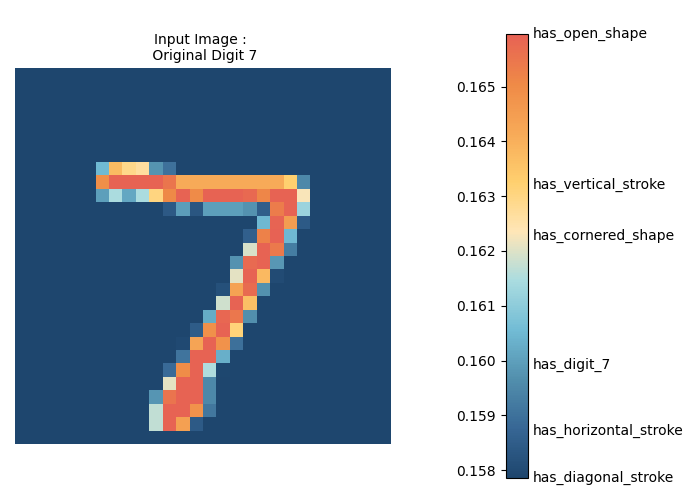}
        \label{fig:1-1}
    }
    \subfigure[Enhanced in corner region.]{
        \includegraphics[width=0.20\textwidth]{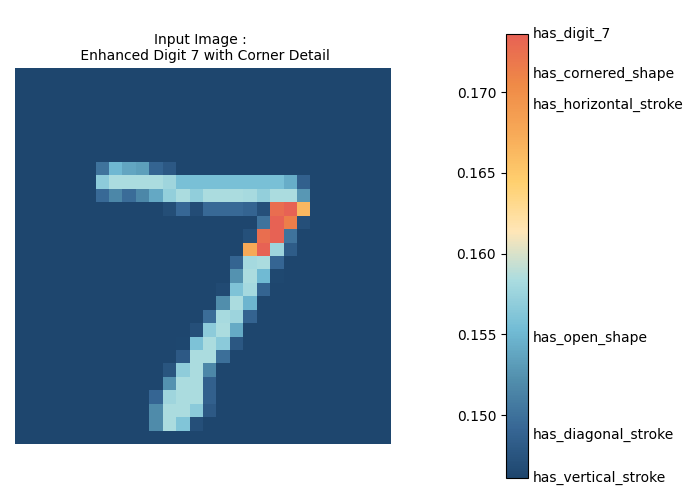}
        \label{fig:1-2}
    }
    \caption{Concept attention (Top 6) of the original/enhanced digit 7 on CMNIST. Enhancing the corner region increases the attention scores of \emph{has corner shape} and \emph{has digit 7}.}
    \label{fig:1}
\end{figure}

This section aims to investigate encoded semantics for individual filters via filter-wise concept attentions. We begin with validating the effectiveness of using learning images to selectively enhance concept representations. Figure~\ref{fig:1} shows an example from CMNIST, where we enhance the corner region by increasing its pixel intensity. As a result, attention scores for the concepts \emph{has digit 7} and \emph{has cornered shape} increase, aligning with the visual modification. Moreover, it suggests that enhancing specific low-level features can facilitate the activation of related high-level concepts -- for example, the cornered shape aids the identification of digit 7.

\begin{figure}[htbp]
    \centering
    \subfigure[Original.]{
        \includegraphics[width=0.19\textwidth]{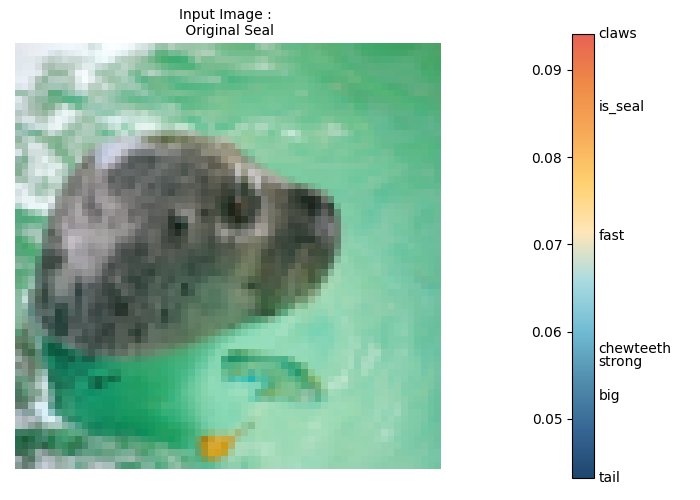}
        \label{fig:2-1}
    }
    \subfigure[Obscured.]{
        \includegraphics[width=0.19\textwidth]{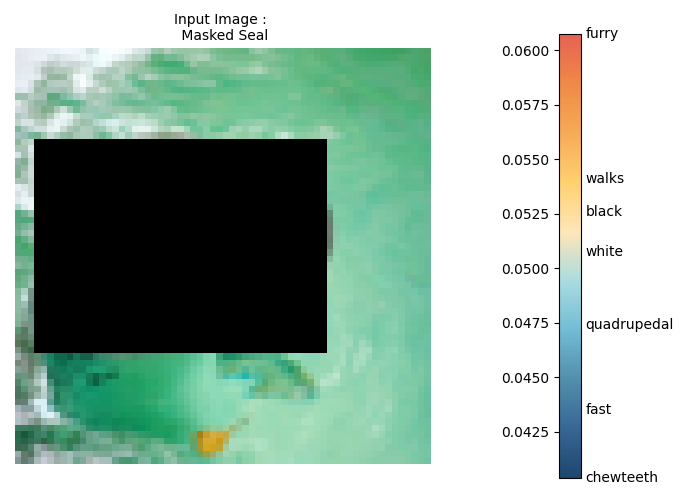}
        \label{fig:2-2}
    }
    \caption{Concept attention (Top 7) of the original/obscured seal input on CAWA. Masking the main object decreases the attention scores of seal-related concepts.}
    \label{fig:2}
\end{figure}

We conduct interventions on CAWA dataset by masking the main object in the image. As shown in Figure~\ref{fig:2}, this leads to a significant drop in the attention for the high-level concept \emph{is seal}, demonstrating the sensitivity and faithfulness of the concept attentions.

\begin{figure}
    \centering
    \includegraphics[width=\linewidth]{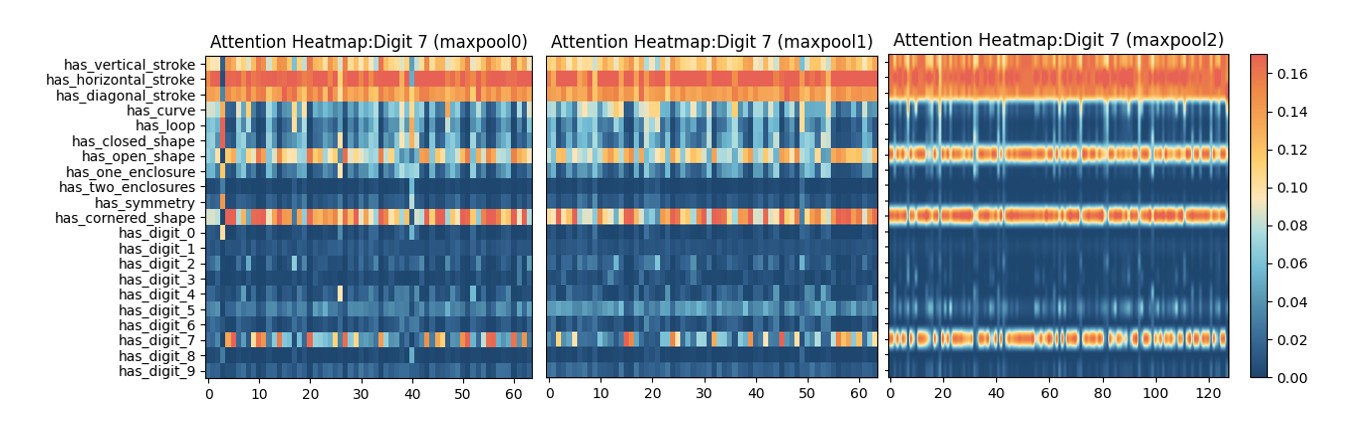}
    \caption{Filter-Wise concept attentions of digit 7 on CMNIST. The horizontal axis represents filter indices, and the vertical axis lists the concept set in hierarchical order. In deeper layers, concept attentions shift toward more class-specific and higher-level concepts.}
    \label{fig:3}
\end{figure}

These observations demonstrate that concept attention reliably reflects the impact of localized visual cues. Therefore, we can highlight the semantic focus of each filter by overlaying their corresponding learning image. Figure~\ref{fig:3} illustrates filter-wise concept attentions for digit 7 on CMNIST dataset. The horizontal axis represents filter indices, and the vertical axis lists the concept set in hierarchical order. In shallower layers, concept attentions are distributed more broadly across low-level concepts. By contrast, the last layer shows sharper focus on class-specific concepts, particularly the high-level concept \emph{has digit 7}.

\begin{figure}
    \centering
    \includegraphics[width=0.7\linewidth]{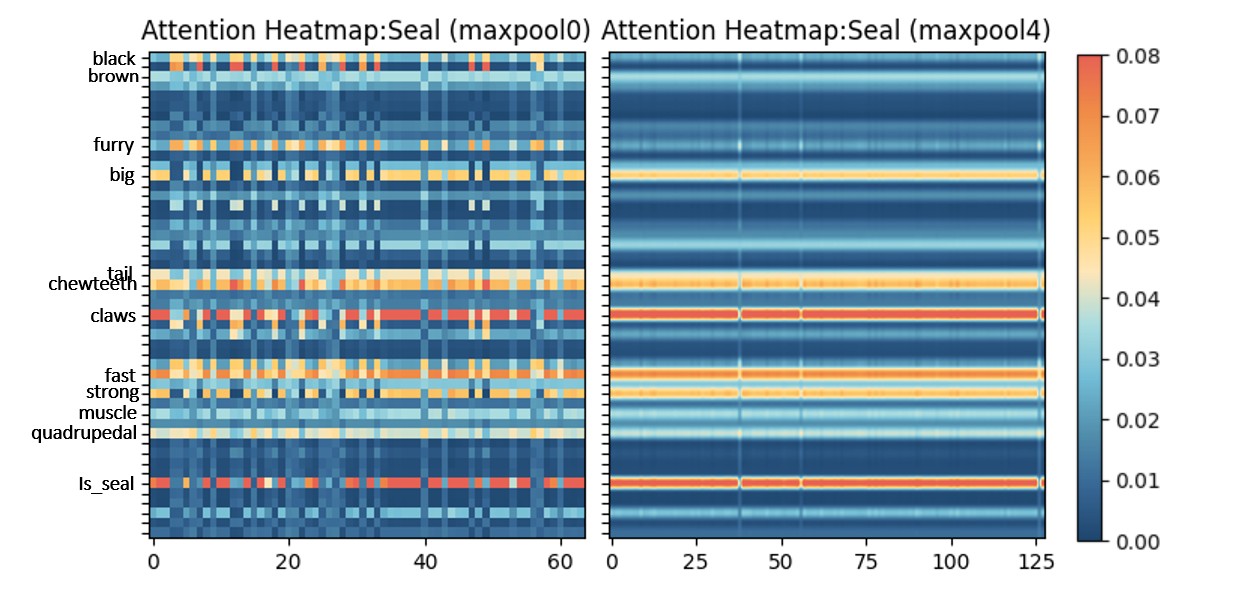}
    \caption{Filter-Wise concept attentions of class seal on CAWA.}
    \label{fig:4}
\end{figure}

A similar trend is observed in Figure~\ref{fig:4}, which presents the concept attentions for the first and last layers of class seal on CAWA. These findings align with the widely accepted understanding that deeper CNN layers tend to capture abstract, class-relevant semantics, while shallower layers primarily respond to general and low-level visual patterns.


\subsection{Tracing Inter-Layer Conceptual Transformations}\label{sec:mid-level} 



Having examined filter-wise concept attentions, we switch to the conceptual transformations across layers. Specifically, we investigate how semantic information propagates and evolves by analyzing inter-layer conceptual pathways.

We first calculate the concept transition matrices between each pair of filters in adjacent layers as described in Section~\ref{method:step2}. Only top $k$ most-activated concepts for each filter are taken into account. The value of $k$ is selected based on empirical validation, which is set to 7/15 for CMNIST/CAWA in this section, and 15/25 in the following section. Detailed analysis is provided in Supplementary Material 4.2.


After obtaining the concept transition matrices between filters, we apply K-means clustering to group similar matrices. Each cluster center serves as a prototypical concept transition matrix, representing a typical semantic pattern between adjacent layers. In this way, we avoid directly analyzing the overwhelming number of inter-filter transitions and focus instead on the inter-layer conceptual interactions.

\begin{figure}
    \centering
    \includegraphics[width=\linewidth]{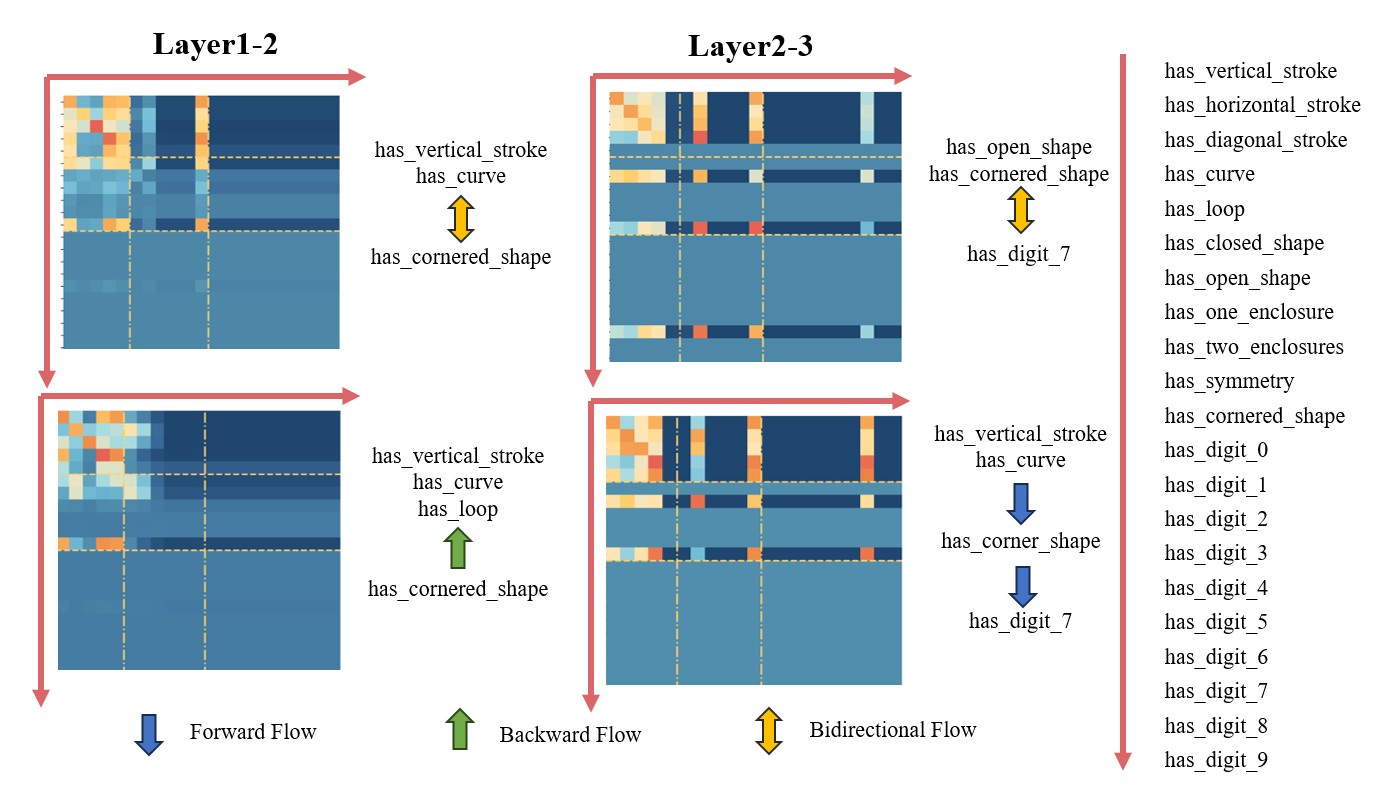}
    \caption{Prototypical concept transition matrices for digit 7 on CMNIST. Blue/green/yellow arrows indicate forward/backward/bidirectional conceptual flows. Red arrow on the right indicates the concept order along both axes. There are some meaningful conceptual pathways, e.g., \emph{has vertical stroke} + \emph{has curve} $\rightarrow$ \emph{has cornered shape} $\rightarrow$ \emph{has digit 7}.}
    \label{fig:5}
\end{figure}

Figure~\ref{fig:5} displays some prototypical concept transition matrices for digit 7 on CMNIST. The blue, green, and yellow arrows indicate forward, backward, and bidirectional conceptual flows respectively. Both the horizontal and vertical axes follow the hierarchical ordering of the concept set, as illustrated with the red arrows on the right. The light dashed lines split the concepts into 3 semantic levels.

The visualization results reveal semantically meaningful transitions. For example, the concepts \emph{has vertical stroke} and \emph{has curve} jointly flow to \emph{has cornered shape}, which subsequently leads to the high-level concept \emph{has digit 7}. This hierarchical progression mirrors how humans recognize digits by composing primitive shapes. Notably, we observe that between layers 1 and 2, the flows primarily occur among low- and mid-level concepts, reflecting early-stage structural composition. In contrast, transitions from layer 2 to layer 3 increasingly involve high-level concepts, suggesting the emergence of abstract, class-specific representations.

\begin{figure}[htbp]
    \centering
    \subfigure[Digit 7 (CMNIST).]{
        \includegraphics[width=0.22\textwidth]{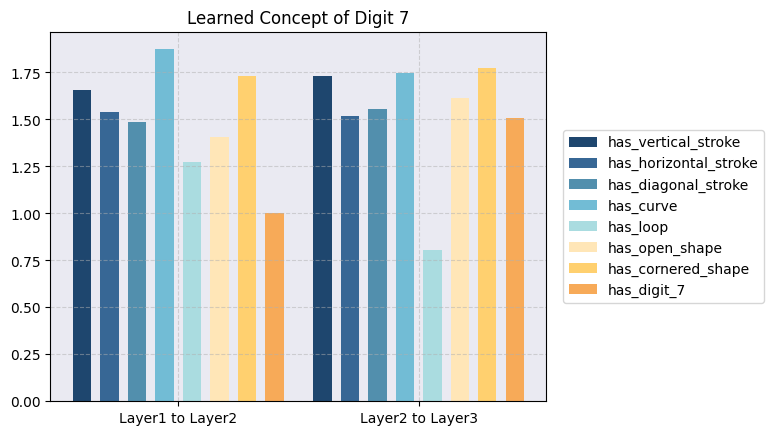}
        \label{fig:8-1}
    }
    \subfigure[Class seal (CAWA).]{
        \includegraphics[width=0.20\textwidth]{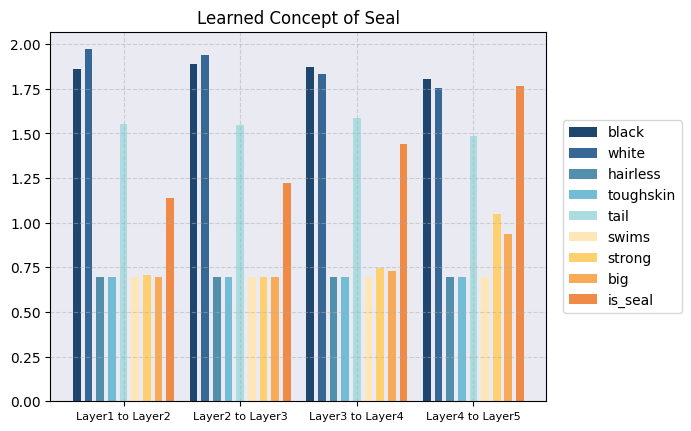}
        \label{fig:8-2}
    }
    \caption{Average concept transition mass. The transition mass concentrates increasingly on high-level concepts across deeper layers.}
    \label{fig:8}
\end{figure}

To quantify the strength and directionality of conceptual transitions, we compute the average concept transition mass across layers. Figure~\ref{fig:8-1} presents the average transition mass of the most relevant concepts for digit 7, with concepts arranged hierarchically from left to right. As the network deepens, we observe a clear increase in transition mass directed toward mid- and high-level concepts. In contrast, lower-level concepts receive diminishing transition mass, indicating that deeper layers progressively consolidate semantic information into more abstract representations.

\begin{figure}
    \centering
    \includegraphics[width=\linewidth]{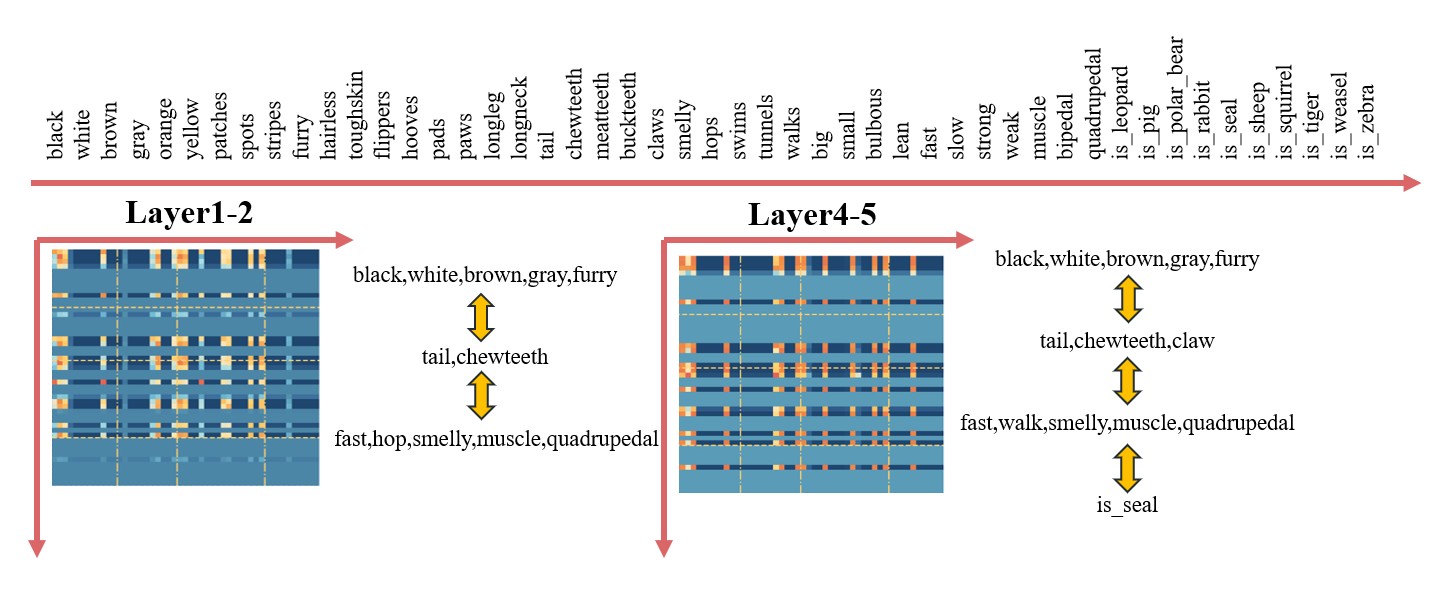}
    \caption{Prototypical concept transition matrices for the class seal on CAWA.}.
    \label{fig:7}
\end{figure}

We replicate the analysis on CAWA and observe a similar hierarchical progression. Figure~\ref{fig:7} shows the prototypical concept transition matrices for the seal class. Low-level concepts such as \emph{black} and \emph{furry} flow into body parts (e.g., \emph{tail}, \emph{claw}), which then integrate into higher-level physical traits (\emph{muscle}) and life habits (\emph{fast}, \emph{walk}) and finally into animal identity (\emph{is seal}). Some unexpected concepts also emerge in the conceptual pathways -- for example, \emph{quadrupedal}. While seals are not typical quadrupeds on land, they have four limbs and often adopt body postures resembling terrestrial quadrupedal animals. This visual similarity likely explains the high relevance of the \emph{quadrupedal} concept.

We also plot the average transition mass for CAWA in Figure~\ref{fig:8-2}, again revealing a progressive concentration on high-level concepts in deeper layers. These findings demonstrate that CNNs gradually build hierarchical semantic representations, and our conceptual pathway analysis effectively uncovers these semantic trajectories, offering an interpretable and human-aligned view of the internal reasoning process. Additional examples and visualizations are provided in Supplementary Material 4.3.

\subsection{Assessing the Impact of Conceptual Pathways on Model Performance}\label{sec:global-level} 



Having analyzed the inter-layer conceptual pathways, we now assess their global impact on model performance. Conceptual pathways reflect the progressive integration of low-level visual features into high-level, class-specific concepts. Intuitively, disrupting a valid conceptual pathway can degrade the model’s ability to make accurate predictions.

To evaluate the functional importance of conceptual pathways, we introduce conceptual pathway pruning. Specifically, if a conceptual pathway exists between filter $F^i_l$ and filter $F^j_{l+1}$, we simulate its removal by zeroing out the $(j,i)$-th entry in the corresponding weight matrix. As discussed previously, a threshold $\tau$ is used to filter out conceptual flows. By gradually decreasing $\tau$, more connections meet the criterion, resulting in a more aggressively pruned network. The same threshold is applied across all layers. Conversely, we can conduct non-conceptual pathway pruning. We compare conceptual and non-conceptual pathway pruning with classical L1-pruning~\cite{liu_learning_2017}, which removes connections based solely on the magnitude of their parameters. In the main text, we conduct global pruning across all layers. Additional layer-wise pruning analysis can be found in Supplementary Material Section 4.4.

Figure~\ref{fig:9-1} presents the performance of the three methods on CMNIST, with the x- and y-axis indicating the pruning ratio and accuracy respectively. At low pruning ratios, conceptual pathway pruning causes minimal accuracy loss, consistent with the observation that many pathways have redundant backups conveying similar semantics. However, as the pruning ratio approaches 30\%, accuracy drops sharply to around 50\%, equivalent to random guessing between even and odd digits. In contrast, L1-pruning results in a more gradual decline in performance, and non-conceptual pathway pruning exhibits the highest tolerance: even after removing nearly 95\% of such connections, accuracy remains around 60\%. The steep decline in accuracy highlights the critical role of conceptual pathways in semantic reasoning.



We further conduct pruning on CAWA. As illustrated in Figure~\ref{fig:9-2}, the pattern of accuracy decline differs from that on CMNIST. Notably, accuracy drops to approximately 60\% when conducting non-conceptual or L1-norm pruning, close to random guessing for this binary task (6:4 split). Conversely, conceptual pathway pruning, while not causing the steepest accuracy decline, can reduce performance to as low as 40\%. -- even worse than random guessing. This finding suggests that it disrupts essential semantic structures rather than merely removing redundant connections. Unlike other pruning strategies, it directly impairs the model’s ability to generalize and reason semantically, which indicates the critic role of conceptual pathways in model performance.


\begin{figure}[htbp]
    \centering
    \subfigure[CMNIST.]{
        \includegraphics[width=0.18\textwidth]{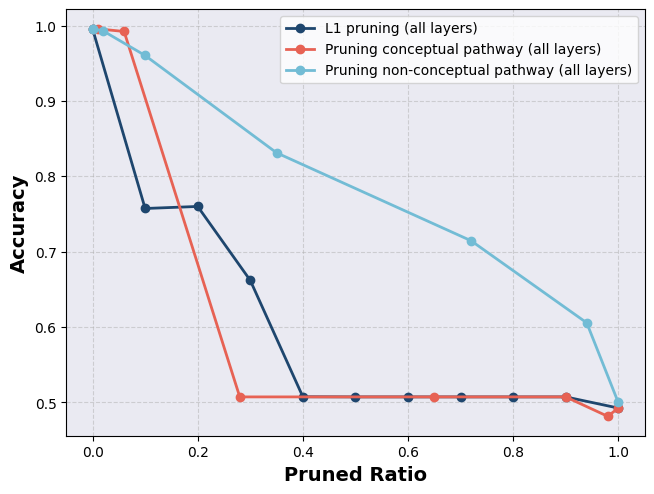}
        \label{fig:9-1}
    }
    \subfigure[CAWA.]{
        \includegraphics[width=0.18\textwidth]{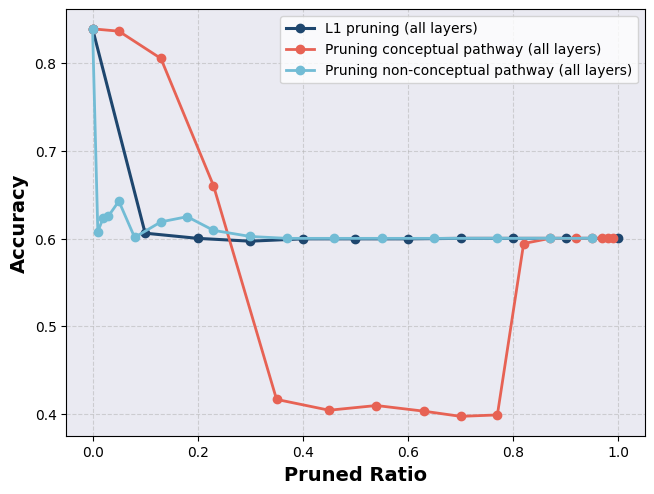}
        \label{fig:9-2}
    }
    \caption{Pruning results on CMNIST and CAWA. For the former, pruning the conceptual pathways leads to the sharpest accuracy drop. For the latter, it causes severe performance degradation, which even falls below random guessing. Those observations highlight the crucial role conceptual pathways play in the model’s decision-making process.}
    \label{fig:9}
\end{figure}

\begin{figure}[htbp]
    \centering
    \subfigure[Number of different pathways.]{
        \includegraphics[width=0.18\textwidth]{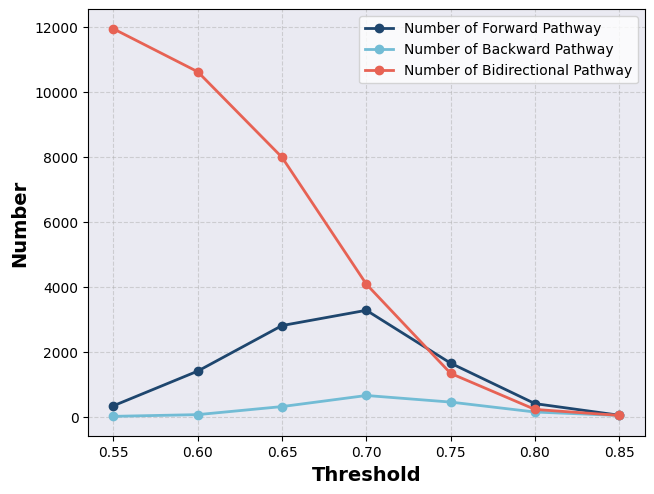}
        \label{fig:10-1}
    }
    \subfigure[Average accuracy drop per pruned pathway.]{
        \includegraphics[width=0.18\textwidth]{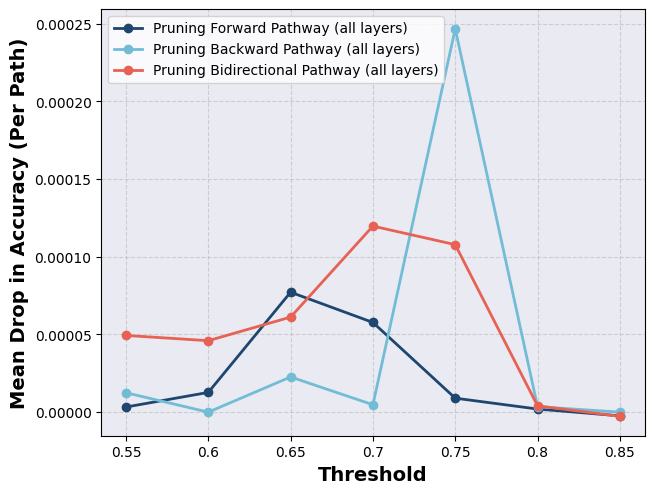}
        \label{fig:10-2}
    }
    \caption{Analysis of different types of conceptual pathways on CMNIST. Bidirectional pathways are the most prevalent, and they exhibit more stable accuracy impacts, suggesting the important role in model performance.}
    \label{fig:10}
\end{figure}

We further categorize pathways into forward, backward, and bidirectional as described in Section~\ref{method:step2}, and analyze their individual contributions on CMNIST. Figure~\ref{fig:10-1} shows the number of each type of pathway extracted under varying thresholds $\tau$. As the threshold increases, the total number of extracted pathways decreases. Among them, bidirectional pathways are the most prevalent, followed by forward and then backward ones. To quantify the importance of each pathway type, we measure the average accuracy drop per pruned pathway, as shown in Figure~\ref{fig:10-2}. As the threshold $\tau$ varies, the curves for all three types exhibit noticeable fluctuations. However, bidirectional pathways demonstrate relatively stable accuracy impacts across different pruning thresholds. These observations imply that conceptual flow is not strictly unidirectional. Instead, low- and high-level concepts frequently interact in both directions. Such bidirectional communication likely facilitates iterative feature refinement and preserves semantic coherence across layers, thus playing a critical role in model performance.


\section{Conclusions}
In this work, we propose ConceptFlow, an interpretability framework for CNNs. It generates filter-wise concept attentions that faithfully reflects the semantics captured by individual filters, and further extracts conceptual pathways that trace how concepts transform and evolve across layers.

We evaluate our method on two self-annotated datasets with hierarchical concepts, demonstrating its ability to provide fine-grained and human-aligned interpretations. Furthermore, we validate the strong correlation between conceptual pathways and model performance, offering new insights into the internal reasoning process of CNNs.



\bigskip

\bibliography{aaai2026}
\end{document}